\documentclass[10pt,letterpaper]{article}

\usepackage{cogsci}

\cogscifinalcopy 

\usepackage{pslatex}
\usepackage{apacite}
\usepackage{float} 

\usepackage{graphicx}
\usepackage{amsmath}
\usepackage{amssymb}
\usepackage{booktabs}
\usepackage[dvipsnames]{xcolor}
\usepackage{lipsum}
\usepackage{multicol}
\usepackage{multirow}
\usepackage{epigraph}
\usepackage[T1]{fontenc}
\usepackage[hyphens]{url}

\DeclareSymbolFont{letters}     {OML}{cmm} {m}{it}
\DeclareMathAlphabet\mathcal{OMS}{cmsy}{m}{n}
\SetMathAlphabet\mathcal{bold}{OMS}{cmsy}{b}{n}

\definecolor{yello}{HTML}{ffb677}
\definecolor{blu}{HTML}{005082}
\definecolor{purpl}{HTML}{726a95}
\definecolor{orang}{HTML}{ff9a76}
\definecolor{tealish}{HTML}{1aa6b7}
\definecolor{greeeen}{HTML}{0e918c}

\newcommand{\km}[1]{#1}

\newcommand{\blank}{$\rule{0.6cm}{0.15mm}$}

\newcommand{\pp}{\mathcal{P}}
\newcommand{\cc}{\mathcal{C}}
\newcommand{\sss}{\mathcal{S}}
\newcommand{\mask}{\textsc{[mask]}}

\newcommand{\induction}[2]{
\begin{equation}
    \begin{tabular}{@{}l@{}}
        #1 \\ \midrule #2
    \end{tabular}%
\end{equation}
}

\makeatletter
\renewcommand{\paragraph}{%
  \@startsection{paragraph}{4}%
  {\z@}{1ex \@plus 1ex \@minus .2ex}{-1em}%
  {\normalfont\normalsize\bfseries}%
}
\makeatother
\title{Do language models learn typicality judgments from text?}
 
\author{
\large \bf Kanishka Misra$^1$ (kmisra@purdue.edu)\\
\large \bf Allyson Ettinger$^2$ (aettinger@uchicago.edu) \\
\large \bf Julia Taylor Rayz$^1$ (jtaylor1@purdue.edu) \\
  $^1$Department of Computer and Information Technology,
  Purdue University, IN, USA \\
  $^2$Department of Linguistics, University of Chicago, IL, USA
}

\begin{document}

\maketitle

\begin{abstract}
Building on research arguing for the possibility of 
conceptual and categorical knowledge acquisition through statistics contained in language, we evaluate predictive language models (LMs)---informed solely by textual input---on a prevalent phenomenon in cognitive science: \textit{typicality}.
Inspired by experiments that involve language processing and show robust typicality effects in humans, we propose two tests for LMs.
Our first test targets whether typicality modulates LM probabilities in assigning taxonomic category memberships to items. 
The second test investigates sensitivities to typicality in LMs' probabilities when extending new information about items to their categories.
Both tests show modest---but not completely absent---correspondence between LMs and humans, suggesting that text-based exposure alone is insufficient to acquire typicality knowledge.

\end{abstract}

\noindent
{\colorbox{yellow}{\textbf{Disclaimer:}}} This is a preprint of an accepted manuscript, which will be published in \textit{Proceedings of the 43rd Annual Conference of the Cognitive Science Society.} (Vienna; held online). The final version is available online at: \texttt{url TBD}.

\section{Introduction}

\noindent
Perhaps one of the most robust findings in the study of human categorical knowledge is the phenomenon of \textit{typicality}, the observation that certain members of a category are considered to be more representative of the category than others \cite{murphy2004big}.
As observed in the seminal work of \citeA{rosch1975cognitive}, native English speakers rate \textit{robins} and \textit{canaries} as more typical birds than \textit{penguins} and \textit{emus}, \textit{chairs} and \textit{sofas} as more typical furniture than \textit{clocks} and \textit{vases}, etc.
Typicality differences in stimuli strongly predict response times in taxonomic sentence verification tasks \cite{rips1973semantic, rosch1973internal} and category production \cite{rosch1976structural}.
In the context of learning, typical items facilitate faster concept acquisition than do atypical items \cite{rosch1976structural}.
Typicality also prominently affects category-based induction \cite{rips1975inductive, osherson1990category}: that is, subjects more readily extend new information about typical---as opposed to atypical---items to the entire category. 
In summary, typicality is a salient and impactful phenomenon in the study of human category knowledge. 

There is a growing body of research on the view that words or language in general act as distributional cues to categorical knowledge, as opposed to mappings onto concepts---that statistics contained in language can, to an extent, inform about the world \cite{lupyan2019words}.
This view is supported by recent evidence in natural language processing (NLP) and cognitive science that shows encouraging signs of computational models learning world \cite{petroni2019language}, categorical \cite{ettinger2020bert}, and conceptual \cite{weir2020probing} knowledge while relying solely on text-based input.
\km{Though these works investigated knowledge of categories (through word prediction-based categorization prompts such as \textit{``A robin is a \blank{} .''}), they do not consider any distinction between central and peripheral members of categories.}

\km{Expanding on the aforementioned promising results related to conceptual and categorical knowledge}, we ask the question: ``How much do the statistical associations contained in text reflect typicality effects in categories?''
To this end, we present a case-study on language models (LMs) that are pre-trained on massive amounts of text and learn representations that are optimized to reflect the statistics of the language used in textual-form. 
We investigate whether the phenomenon of typicality emerges as a result of LM pre-training.
Our tests are grounded in the psychological study of concepts and categories, and are inspired by prior human experiments that show clear sensitivities to typicality in processing of textual stimuli.
First, we build on prior work analyzing conceptual and categorical knowledge in LMs, and test whether typicality effects modulate LM judgments of taxonomic sentence verification (\textit{``a robin is a bird''}) as they do in humans \cite{rips1973semantic, rosch1973internal}.
Complementing this simple and direct test of taxonomic category membership, we add a layer of complexity, and investigate the manifestation of typicality effects in LMs on the basis of how they extend new information about items (\textit{``robins can dax''}) to all members of a category (\textit{``all birds can dax''}), inspired by tests targeting psychological strength of inductive arguments  \cite{osherson1990category}.
Though the human experiments that inspire our tests do not explicitly target typicality as a phenomenon, typicality effects still robustly modulate human behavior on them.
Hence, we examine whether LMs show comparable typicality effects on stimuli similar to those used in the above experiments.

We find non-trivially positive but modest sensitivities of LMs to typicality effects in both our experiments.
We also find LMs, on average, to be less extreme in their sensitivities to atypical and typical items as compared to humans.
This suggests that the word prediction capacities of LMs that are optimized to reflect the statistics that are contained in textual corpora are moderately influenced by typicality effects in assessing strength of simple taxonomic verification as well as more complex inductive inferences about categories.
Our results reflect the difficulty of acquiring human-like category knowledge without extra-linguistic input, at least with the current computational models of language processing.

\section{Materials and Methods}
\subsection{Models Studied}
We conduct our analyses on pre-trained LMs based on the transformer architecture \cite{vaswani2017attention}.
Our choice of LMs is motivated by recent evidence that shows qualitative alignment of category knowledge (\textit{``a robin is a \underline{bird}''}, \textit{``a \underline{bear} has fur, has claws.''}) in pre-trained LMs \cite{ettinger2020bert, weir2020probing}.
Although we focus on a particular type of pre-trained LMs (transformers) in this paper, the tests we propose can be applied to any LM.
We investigate two broad classes of transformer-based pre-trained LMs: \textbf{(1) Incremental LMs}, trained autoregressively (left to right) to predict one word at a time, when conditioned on exclusively the left context; and \textbf{(2) Masked LMs}, that access context of the word to be predicted bidirectionally, e.g., the models are optimized to predict correct completions (\textit{airplane} or \textit{bird}) to sentences such as \textit{``the \mask{} flew away,''} where \mask{} represents the hidden word.
We apply our tests on GPT \cite{radford2018improving} and GPT2 \cite{radford2019language} as our Incremental LMs, and ALBERT \cite{lan2019albert}, ELECTRA \cite{clark2020electra}, BERT, \cite{devlinBERTPretrainingDeep2019} and RoBERTa \cite{liu2019roberta} as our Masked LMs.
In addition, we use compressed versions of the above models \cite{sanh2019distilbert}: distilGPT2, distilBERT-base, and distilRoBERTa-base.
All transformer-based pre-trained LMs were accessed using the \texttt{transformers} library \cite{wolf-etal-2020-transformers}.

Finally, we also used a 5-gram language model with kneyser-ney smoothing, trained using the \texttt{KenLM} toolkit \cite{heafield2011kenlm}, as a baseline model that lacks the kind of representational learning mechanisms that empower the above models.
This model is trained on the Dec, 2020 dump of English Wikipedia.\footnote{\url{https://dumps.wikimedia.org/enwiki/20201220/}} 
The performance of the 5-gram model represents the extent to which our tests can be approximated simply by memorizing sequences of up to 5 words in length.

\subsection{Data and Stimuli}
\paragraph{Item typicality data} 
For both experiments, we use as our primary source the list of 565 item-typicality ratings compiled by \citeA{rosch1975cognitive} across 10 different categories.
In the original human experiments, 209 native speakers of English were tasked to rate the ``goodness of example'' for various items of each given category, on a scale of 1 (most typical) to 7 (least typical).
The statistics of the items and categories is presented in Table \ref{tab:categoryitems}.
It should be noted that the experiments we base our tests on involve sensitivities to typicality measured using different quantities (response times and raw typicality ratings), but make none or only a small subset of results available. 
Therefore, we use the \citeA{rosch1975cognitive} ratings as the common ``ground-truth'' typicality ratings for our experiments.

\paragraph{Stimuli Setup}
Because the models we investigate are sentence processors, and because all of our tests involve propositions about items and categories expressed as sentences, we rely on using sentence stimuli in our experiments.
Every stimulus consists of two components: (1) condition, which is a noun phrase/sentence consisting of the item (\textit{robins, sparrows, eagles, etc.}); and (2) predicted material, which consists of the super-ordinate category (\textit{bird}). 
\km{The exact linguistic format in which it appears depends on the experiments --- we use single words as the predicted material in our Taxonomic Sentence Verification experiment while for our Category-based induction experiment we use a full sentence as our predicted material.}
In evaluating typicality measurements of various items for a given category, the predicted material remains constant, while the condition changes depending on the item.
\km{Table \ref{tab:example} shows examples of stimuli we use in each of our experiments.}


\begin{table}[t!]
\vspace{-1.5em}
\def\arraystretch{1.15}
\centering
\caption{Number of items (\textit{N}) per category \cite{rosch1975cognitive}.}
\vspace{1em}
\label{tab:categoryitems}
\begin{tabular}{|c|c|c|c|}
\hline
\textbf{Category} & \textbf{\textit{N}} & \textbf{Category} & \textbf{\textit{N}} \\ \hline
furniture         & 60                  & vegetable         & 56                  \\ 
tool              & 60                  & clothing          & 55                  \\ 
toy               & 60                  & bird              & 54                  \\ 
weapon            & 60                  & fruit             & 51                  \\ 
sport             & 59                  & vehicle           & 50                  \\ \hline
\end{tabular}
\vspace{-1em}
\end{table}

\subsection{Measures}
Following precedent set by previous work evaluating conceptual knowledge in pre-trained LMs, we use the models' probability estimates as our main variable of interest.
Specifically, we focus on the log-probability of the word or statement represented in the predicted material, given the condition, $\log p_{\textit{LM}}(\textrm{predicted} \mid \textrm{condition})$, i.e., we are measuring the effect on the probability of the predicted part (held constant for a given category) due to the item mentioned in the condition.
Our reason for separating the item from the predicted material is two-fold:
(1) it avoids skewed measurements due to the choice of determiner 
(\emph{a} vs \emph{an}) that precedes the item in the condition (a model might assign higher value to $p(\textit{ostrich} \mid \textit{an})$ simply due to a component that is sensitive to determiner prefixes), or when the model does not include the item word in its vocabulary,\footnote{E.g., RoBERTa segments the word \textit{ostrich} into \textit{ostr} and \textit{ich}, and during estimation, the probability of \textit{ich} given that it is preceded by \textit{ostr} is anomalously high, skewing the overall sequence probability.} and (2) it aids in factoring out the role played by the frequency of the item in the condition -- the model can prefer an item over the other simply due to its frequency in the training corpus.
While it is straightforward to compute our conditional probability measure for incremental LMs by using the chain-rule, we rely on recent work by \citeA{wang-cho-2019-bert} to approximate sequence log-probabilities in Masked LMs by summing the conditional log-probabilities of all words in the stimuli.
\begin{figure*}[t!]
    \centering
    \includegraphics[width=\textwidth]{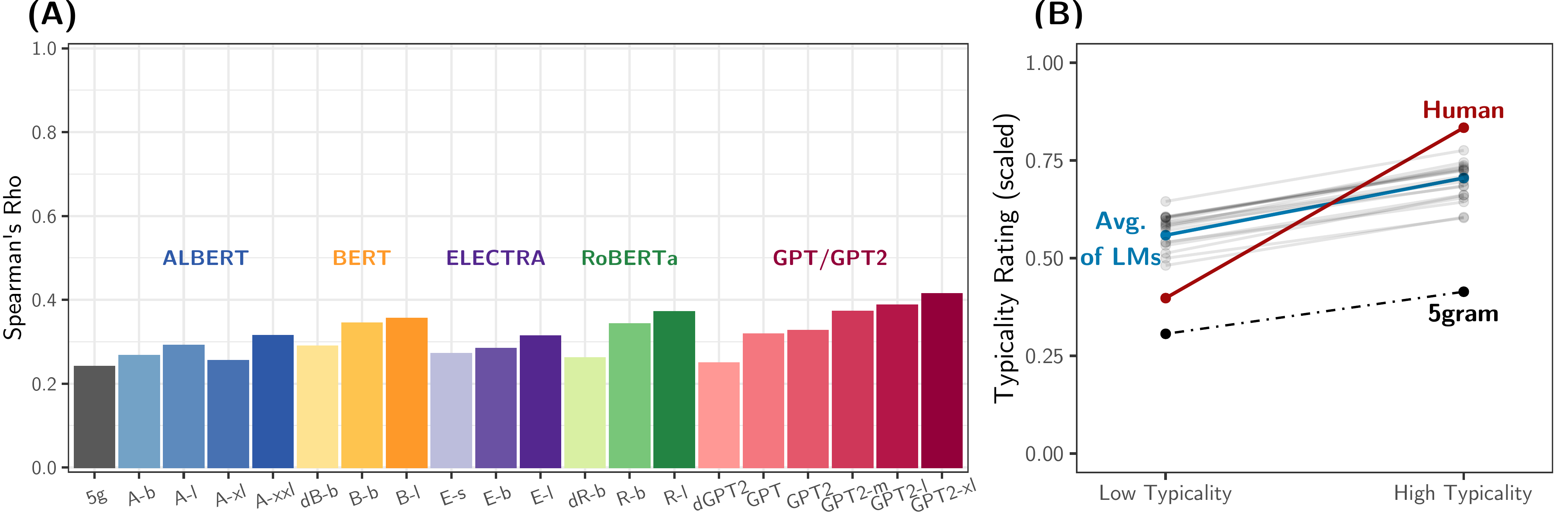}
    \caption{\textbf{(A)} Spearman correlation ($\rho$) measured between LM log-probabilities assigned to word completion in taxonomic stimuli (experiment 1) and typicality ratings from \citeA{rosch1975cognitive}. Models from the same family are arranged in an increasing order of total number of parameters. \textbf{(B)} Scaled typicality scores from LMs (log-probabilities on taxonomic stimuli) and Humans (raw ratings) between low and high typicality category members.}
    \label{fig:tsv}
    \vspace{-1em}
\end{figure*}

\section{Experiments}
\subsection{1. Taxonomic Sentence Verification}
\paragraph{Phenomenon} Typicality effects in the sentence verification paradigm were introduced by \citeA{rips1973semantic} and \citeA{rosch1973internal}.
Subjects were tasked with verifying the truth of sentences expressing taxonomic propositions, such as \textit{``An $X$ is a $Y$''}---where $X$ and $Y$ are the item and category, respectively.
The subjects consistently responded faster to verifying the truth of propositions where $X$ was a typical member of $Y$ than when it was an atypical one. 
\paragraph{Linking Phenomenon to LMs} We draw on the aforementioned findings and investigate whether typicality is able to account for difference in the word probabilities to complete taxonomic sentences by our tested LMs.
Linking our hypothesis to the original experiment requires a simplifying assumption that an LM's sequence log-probability is proportional to its plausibility for a sequence.
\km{That is, we assume and expect a semantically sound LM to show overall high probability scores for semantically plausible propositions, which in this case, are simple taxonomic propositions\footnote{However, we acknowledge that this might not always be the case. For instance, LMs are largely insensitive to negation and semantic role-reversal \cite{ettinger2020bert}.}.
Therefore, LMs that are more sensitive to typicality effects should show greater magnitudes for the measure $\log p_{\textit{LM}}(Y \mid \textit{An }X\textit{ is a})$ when $X$ is a more typical member of $Y$.}

\begin{table}[t!]
\def\arraystretch{1.5}
\centering
\caption{Examples of stimuli used in our experiments. Our measures take the form: $\log p(\textrm{predicted} \mid \textcolor{blu}{\textrm{condition}})$}
\vspace{1em}
\label{tab:example}
\begin{tabular}{|p{0.35\linewidth}|p{0.54\linewidth}|}
\hline
\textbf{Experiment} & \textbf{Stimulus} \\ \hline
Taxonomic Sentence Verification & $\textcolor{blu}{\underbrace{\textrm{A robin is a}}_\textrm{condition}}\underbrace{\textrm{bird}}_\textrm{predicted}$. \\ \hline
Category-based Induction & $\textcolor{blu}{\underbrace{\textrm{Saws can dax.}}_\textrm{condition}}\underbrace{\textrm{All tools can dax.}}_\textrm{predicted}$ \\ \hline
\end{tabular}
\vspace{-1em}
\end{table}

\paragraph{Experiment}
We follow \citeA{rips1973semantic} and \citeA{rosch1973internal} and construct sentences expressing taxonomic propositions using items from the \citeA{rosch1975cognitive} data, i.e., \textit{``An $X$ is a $Y$,''} amounting to 565 unique propositions.
We test for typicality effects by measuring the Spearman correlation ($\rho$) of the sequence log-probability $\log p_{\textit{LM}}(Y \mid \textit{An }X\textit{ is a})$ with the human typicality ratings for items, as collected by \citeA{rosch1975cognitive}.
This correlation measure reflects the extent to which the predictive estimates of an LM reflect typicality information---or information that underlies it---to assess taxonomic verification in sentences.
Additionally, we perform a median split on the \citeA{rosch1975cognitive} ratings by the items' typicality ratings, per category, leaving us with two sets of typical and atypical ratings.
We then compute the average log-probabilities assigned to items in each set and compare them to the average ratings elicited by humans.
All scores in this analysis are re-scaled to be between 0 and 1.

\paragraph{Results}
Figures \ref{fig:tsv}A and \ref{fig:tsv}B show results from our correlation and typicality-effect comparisons.
Non-trivially positive but modest correlations between LM log-probabilities and human typicality ratings ($\rho \in$ [0.24, 0.41], $p < $ .001) suggest that LMs' judgments of taxonomic propositions are moderately reflective of typicality effects. 
Though all LMs assign greater probability scores to category items with high---as compared to low---typicality (see Figure \ref{fig:tsv}B), they are consistently less extreme as compared to humans ($p <$ .001 across all models).
Correlation of 5-gram LM log-probabilities, though weakest in magnitude, are highly competitive with certain smaller yet highly expressive LMs (ALBERT-b, ALBERT-xl, distilGPT2, and distilRoBERTa).
This suggests that a substantial portion of the observed correspondence between model and human typicality judgments can be attributed to fairly simpler sequential statistical effects in word prediction (e.g. memorizing n-grams).
Interestingly, with the exception of the ALBERT family of Masked LMs, models with greater number of parameters tend to show greater correspondence with humans on taxonomic judgments ($\rho = 0.82, p < .001$), suggesting that the information needed to distinguish typical vs. atypical category members during taxonomic attribution requires greater model expressivity.

\subsection{2. Category-based Induction}
\paragraph{Phenomenon}
Typicality of items plays a salient role in making inductive inferences about categories \cite{rips1975inductive}\km{, i.e., when informed about a member $m$ of a category $c$ having a novel property $\gamma$, people are more likely to extend the presence of $\gamma$ to all members of $c$ when $m$ is typical or central to $c$.}
This was more robustly illustrated by \citeA{osherson1990category} in their study exploring the psychological strength of categorical inductive arguments. 
An argument is a finite set of sentences of the form $\pp_1, \pp_2, ..., \pp_n/\cc$, where $\pp_1, \pp_2, ..., \pp_n$ are the argument's premises and $\cc$ is its conclusion.
In categorical arguments, $\pp$ and $\cc$ take the form ``\textit{All members of \textsc{cat} have property $\gamma$},'' where \textsc{cat} is a natural category such as \textit{car} or \textit{sofa}, and the property $\gamma$ remains constant across $\pp$ and $\cc$.
Such arguments can be visualized by separating the premises and conclusions by a horizontal line, like in (1) and (2).
The psychological strength of inductive arguments, for a subject $\sss$, is the degree to which $\sss$'s belief in $\pp$ strengthens their belief in $\cc$ \cite{osherson1990category}.
\induction{Robins have property $\gamma$.}{All birds have property $\gamma$.}
\vspace{-0.5em}
\induction{Penguins have property $\gamma$.}{All birds have property $\gamma$.}
\noindent
\km{Unlike deductive arguments, which involve logical reasoning, inductive arguments such as (1) and (2) involve probabilistic reasoning, i.e., there is an epistemic uncertainty whether the conclusions follow from the given premise (for a detailed review, see \citeNP{feeney2007inductive}).}
A caveat in the Osherson et al. experiments is that the property space $\Gamma = \{\gamma_1, ..., \gamma_n\}$ only includes properties that are unfamiliar to $\sss$, such that the influence of prior knowledge about the properties on the induction process is minimal.
Such properties are also known as \textit{blank predicates} --- for instance, \citeA{osherson1990category} use properties such as \textit{love onions, have sesamoid bones, etc.}

Typicality effects are one of the 13 phenomena examined by \citeA{osherson1990category}.
Specifically, for single-premise arguments where the category of the conclusion subsumes that of the premise, subjects were more likely to believe in the conclusion when the category of the premise was a more typical member of the category in the conclusion, i.e., the argument strength of (1) was found to be greater than that of (2) since \textit{robins} are more typical birds as compared to \textit{penguins}.

\paragraph{Linking Phenomenon to LMs} The \citeA{osherson1990category} study explicitly targets the degree to which uncertain statements such as ``all birds love onions'' are judged in light of new information about a subordinate category such as \textit{robins}.
Analogously, we are interested in assessing whether sophisticated LMs show similar behavior in assigning probabilities to conclusions when conditioned on premises whose categories vary based on their typicality.
If LMs show sensitivity to the typicality of items in this setting, i.e., their log-probability is greater for conclusions with typical versus atypical premise, then we take this as the extent to which typicality---or the factors that underlie it---modulates inductive inference in LMs.
We formulate an approximation of inductive argument strength ($AS$) in an LM as the probability it assigns to the conclusion when conditioned on a given premise. 
For instance $AS(\textit{robin}, \textit{bird})$ for the property \textit{``love onions''} is given by:
\begin{gather*}
    \log p_{\textit{LM}}(\textit{``All birds love onions.''} \mid \textit{``Robins love onions.''})
\end{gather*}
The premise and conclusion sentences naturally fit within our stimulus setup discussed earlier --- the premise sentence is the condition, and the conclusion sentence the predicted material.

\begin{figure*}[t!]
    \centering
    \includegraphics[width = \textwidth]{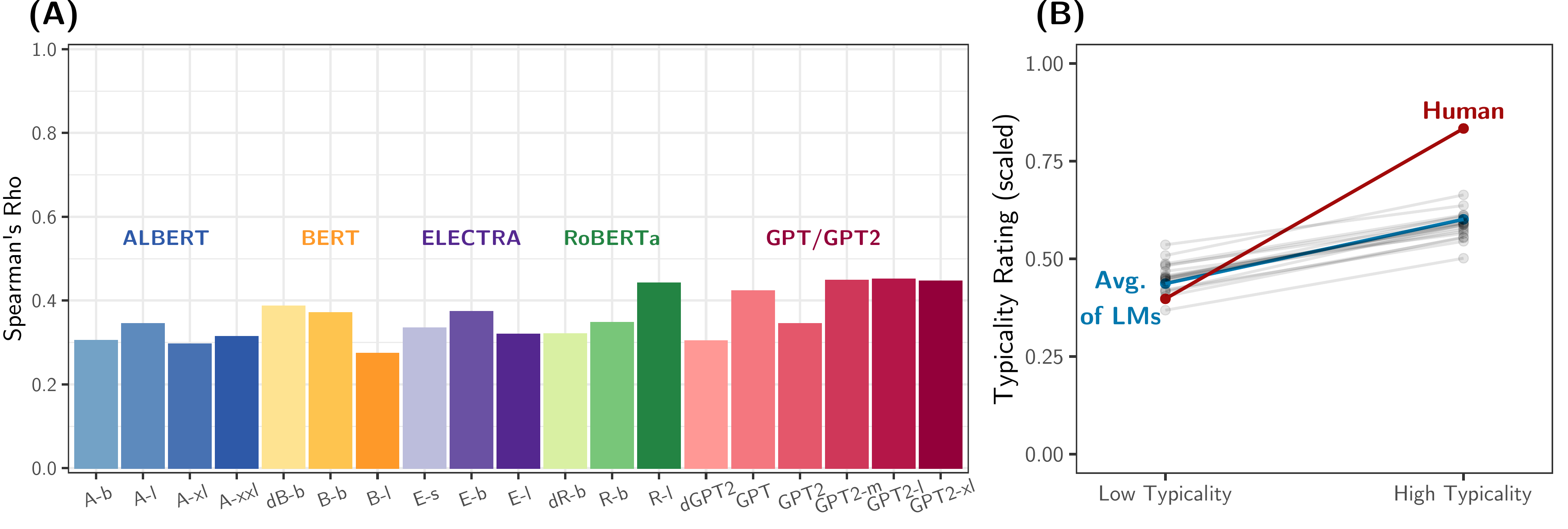}
    \caption{\textbf{(A)} Spearman correlation ($\rho$) measured between average $AS$ scores and human typicality ratings compiled by \citeA{rosch1975cognitive}. Models from the same family are arranged in an increasing order of total number of parameters. \textbf{(B)} Scaled typicality scores from LMs ($AS$ values) and Humans (raw ratings) between low and high typicality category members.}
    \label{fig:induction}
    \vspace{-1em}
\end{figure*}

\paragraph{Experiment}
For our items and categories we use the same stimuli from the previous experiment.
Since \citeauthor{osherson1990category} do not make all of their blank predicates 
available, we construct synthetic properties using nonce words such as \textit{dax}, \textit{wugs}, \textit{feps}, \textit{vorpal}, etc., such that these words do not occur in the vocabulary\footnote{\km{Due to their tokenization mechanism, the LMs we study are always able to encode any text through `word pieces' instead of relying on \texttt{\textless unk\textgreater} tokens.}} of the models, conforming to the blank predicate condition applied by \citeA{rips1975inductive} and \citeA{osherson1990category}.
We create between 15 to 30 properties\footnote{The choice of properties depends largely on the class of word the items belong to, such that syntactic constraints are met. For instance, if \textit{dax} is a verb, it would be ungrammatical to have ``can dax'' as a property of sports, which can be better paired with properties such as ``involve'' and ``require''. The entire unique set of synthetic properties and our construction method is made available in our supplementary materials.} for all items in each category, resulting in a total of 12,180 premise-conclusion pairs across 10 categories.
An example \km{of the stimuli we use for our category-based induction task is shown in Table \ref{tab:example}.
We calculate the $AS$ metric for each premise-conclusion pair with each of our tested LMs.}

Conditioning our LMs as we do here has two potential confounds: 
\textbf{(1) Premise Order Sensitivity (POS)}: A model might estimate high probabilities for words in the conclusion simply because it is relying on lexical cues in its premise \cite{misra2020exploring}, instead of processing the premise compositionally and making inferences about items possessing a property. 
We account for this confound by also computing the LMs' average probability for the conclusion sentence when prefixed by a shuffled version of the premise (10 times, with random seeds).
\textbf{(2) Taxonomic Sensitivity (TS)}: LMs might tend to repeat the property phrase mentioned in the predicted material with high probability when prefixed by a sentence containing it, i.e., repeating \textit{``can dax''} in the conclusion when already conditioned on the same phrase in the premise, confounding the degree to which the conclusion is generated using the taxonomic relationship between the premise and the conclusion categories.
To account for this tendency, we compute the LMs' probabilities for conclusions consisting of a different category with the exact same property as the original (for instance, \textit{``All fruits are slithy''} given \textit{``Sofas are slithy''}).

We find that a substantial amount of variance in our original $AS$ scores is in fact captured by both these confounds (overall $r^2$ = 0.43, $\beta_{\textsc{ts}} = 0.68, \beta_{\textsc{pos}} = -0.04, p < .0001$ in both cases).
We regress these relationships out from our $AS$ scores by first fitting a multiple regression model to predict $AS$ using our confounds, and then subtracting the relationship with TS and POS as follows:
\begin{align*}
    AS  &= \beta_0 + \beta_1\textrm{TS} + \beta_2\textrm{POS} + \epsilon\\
    AS^\prime &= AS - \beta_1\textrm{TS} - \beta_2\textrm{POS}\\
    &= \beta_0 + \epsilon\tag*{(Adjusted $AS$)}
\end{align*}
Using the adjusted $AS$ scores in each LM, we compute the score of generating the conclusion (scaled between 0 and 1) for each category, item, and synthetic property, and average them to get the model's overall score for extending new information about an item to its category.
As was the case in our taxonomic verification judgement, we compute the correlation between our normalized adjusted $AS$ scores and the human typicality ratings from \citeA{rosch1975cognitive}, and compare average $AS$ scores (across all blank properties that we used in this experiment) and human typicality ratings assigned to low and high-typicality items.
Note that since the 5-gram LM predicts word probabilities by conditioning only up to four preceding tokens, which is far fewer than the number of tokens in our stimuli, it shows constant $AS$ values in this experiment.

\paragraph{Results} Figure \ref{fig:induction} summarizes results from our induction experiments.
When LMs extend information about an item to its category, they are moderately but positively influenced by its typicality ($\rho \in$ [0.27, 0.45], $p <$ .001).
This influence is above and beyond their usual predilection towards repeating sequences and being lexically sensitive to items present in the premise \cite{misra2020exploring}.
Deviating from results in the previous experiment, we observe Incremental LMs to show stronger correspondence with human ratings as compared to Masked LMs of comparable size, suggesting that they are slightly more sensitive to the typicality of the premise item in generating the conclusion.
Unlike the previous experiment, we notice almost no effect of model size (in terms of parameters) on the results, suggesting that while making typicality-sensitive attribution of items to their super-ordinate categories is generally improved by scaling up the overall expressiveness of the model, the factors that underlie typicality effects in category-based induction are likely independent of the number of parameters of an LM.

\section{General Discussion and Conclusion}
Extensive research in the field of cognitive science has highlighted the prevalent role played by typicality in studies of categories---that certain items (\textit{chair}) are considered to be better representatives of a category (\textit{furniture}) than others (\textit{vase}).
Motivated by recent evidence showing pre-trained LMs to capture patterns exhibiting conceptual and categorical knowledge, we presented two experiments targeting sensitivities to typicality in LMs.
The first experiment targets typicality directly, in its role played in associating items to their taxonomic categories (\textit{``football is a \underline{sport}''}).
Our second experiment complements this by instead assessing the extent to which the ``knowledge'' of category typicality is used to extend information about items (\textit{``football involves blicking''}) to their respective categories (\textit{``all sports involve blicking''}).
We investigate typicality effects in LMs by evaluating their log-probabilities in response to stimuli as measures of (1) taxonomic verification and (2) inductive argument strength (when conditioned on a premise).
For each test, we made the simplifying assumption that the likelihood assigned by the LM to the sentence stimuli corresponds to the variables of interest---strength of category membership in the first experiment, and argument strength in the second.
Overall, the pre-trained LMs showed positive but modest correlations with human typicality ratings in both experiments, and were, on average, far less extreme in distinguishing between typical and atypical items than humans.
We also observed that a considerable amount of sensitivity to typicality effects can be attributed to the mechanisms available to simpler LMs (5-gram), relative to the sophisticated pre-trained LMs that we studied here, suggesting that the representational mechanisms in most models that are optimized to reflect the statistics in training corpora only account for a minimal gain over correspondence that is afforded by simpler sequential statistics.
Results on pre-trained LMs suggests that the statistical associations that inform their word probabilities are modestly sensitive to human-elicited typicality ratings in (1) attributing items to their category members, as well as (2) making complex inductive inferences about categories when conditioned on new information about the items.
While our taxonomic sentence verification experiments showed typicality correspondence to increase with model size, this was not the case in our induction experiments, suggesting that extending new information about items to their categories in a manner that is positively modulated by typicality effects does not scale with an increase in parameters.
We leave fine-grained exploration of specific language modelling factors affecting typicality correspondence for future work.

LMs are trained by exclusively relying on distributional evidence to inform their word predictions.
In our experiments, we find that while the aforementioned word prediction capacities show qualitatively similar patterns of associating concepts with human-produced property norms \cite{weir2020probing}, they show weak agreement with the typicality effects that are robustly elicited in humans \cite[and references therein]{murphy2004big}.
This suggests that solely relying on text is insufficient for exhibiting quantitatively similar categorical knowledge to that in humans\km{, and highlights the limitations of using word-prediction capacities from state-of-the-art pre-trained LMs as mechanisms to model semantic cognition.}
This is in line with work in knowledge acquisition through text, which suggests large textual corpora to lack real world grounding, in that these corpora represent language use but distort general knowledge about the world \cite{gordon2013reporting}.
Even though text data contain encyclopedic knowledge, they miss out on the more perceptual or semi-perceptual features that can be learned through visual input, and that have been found to better align with human ratings of typicality, albeit on non-taxonomic categories \cite{lake2015deep}.
Another line of work supporting the lack of typicality signal in textual corpora is that of \citeA{bergey2020children}. 
These authors analyze parent-child interactions using models that are similar to---but less-sophisticated than---pre-trained LMs, and find them to negatively align with typicality ratings on adjective-noun compounds. The authors conclude from their findings that much of what children hear (corresponding to language use by the parent) is atypical, as opposed to typical information about noun concepts (specifically with respect to the adjectives that modify them).
While our results also shed light on the difficulty of acquiring knowledge about typical members of categories, they do suggest the presence of some typicality effects, by contrast to the findings of \citeA{bergey2020children}---raising the possibility that associations in text that impact typicality of adjective-noun compounds could be independent of, or even run in opposition to, those that impact taxonomic categories.
At the same time, considering that we do see non-zero correspondence with human typicality ratings, our results also suggest that textual corpora are not fully devoid of associations that may align with empirical phenomena underlying typicality effects.
Taking this as inspiration, future work on modeling of typicality through text will likely require models to correct for the distorted frequency of atypical items mentioned in text, and potentially also include features informed from a more grounded source of knowledge.
\km{One promising way of doing so could be to let LMs and their representations adapt to texts represented as more explicit sources of concept and categorical knowledge \cite{bhatia2020transformer} --- potentially in the form of statements such as \textit{a robin has wings}.
Explicitly encoding features into LMs could possibly make them compliant with feature-based hypotheses of typicality \cite{rosch1976structural} and inductive reasoning \cite{sloman1993feature}, and better facilitate research into other key facets of semantic cognition \cite{rogers2004semantic} in models that learn through text.}

\paragraph{Acknowledgments} We thank the three anonymous reviewers and the meta-reviewer for their comments and feedback. This research has benefited from fruitful discussions with the members of the AKRaNLU lab at Purdue University, and the CompLing lab at the University of Chicago. We also thank the Department of CIT at Purdue University for providing hardware resources for computation.

\paragraph{Reproducibility} To facilitate further research into manifestation of typicality in language processing models, we make our code and supplementary materials available at: \url{https://github.com/kanishkamisra/typicalityprobing}


\bibliographystyle{apacite}

\setlength{\bibleftmargin}{.125in}
\setlength{\bibindent}{-\bibleftmargin}

\bibliography{CogSci_Template}

\end{document}